\documentclass{article}

\usepackage{microtype}
\usepackage{graphicx}
\usepackage{subfigure}
\usepackage{booktabs} 

\usepackage{hyperref}

\usepackage[accepted]{icml2024}

\usepackage{amsmath}
\usepackage{amssymb}
\usepackage{mathtools}
\usepackage{amsthm}

\usepackage[capitalize,noabbrev]{cleveref}

\usepackage{wrapfig}
\usepackage{bbding}
\usepackage{amsmath}
\usepackage{bbm}
\usepackage{array}
\usepackage{tablefootnote}
\usepackage{multirow}
\usepackage{colortbl}
\usepackage{ulem}
\usepackage{CJKutf8}

\usepackage{epigraph}
\newcommand{\prompt}[1]{\textcolor{black}{\small{\texttt{#1}}}}
\newcommand{\g}[1]{\textcolor{gray}{#1}}

\definecolor{Tabcolor}{rgb}{0.99, 0.96, 0.94}
\definecolor{Tabcolor2}{rgb}{0.99, 0.88, 0.86}

\theoremstyle{plain}

\theoremstyle{definition}

\theoremstyle{remark}

\usepackage[textsize=tiny]{todonotes}

\icmltitlerunning{Libra: Building Decoupled Vision System on Large Language Models}

\begin{document}

\twocolumn[
\icmltitle{Libra: Building Decoupled Vision System on Large Language Models}



\icmlsetsymbol{equal}{*}

\begin{icmlauthorlist}
\icmlauthor{Yifan Xu}{casia,pcl,ucas}
\icmlauthor{Xiaoshan Yang}{casia,pcl,ucas}
\icmlauthor{Yaguang Song}{pcl}
\icmlauthor{Changsheng Xu}{casia,pcl,ucas}
\end{icmlauthorlist}

\icmlaffiliation{casia}{MAIS, Institute of Automation, Chinese Academy of Sciences}
\icmlaffiliation{ucas}{School of Artificial Intelligence, University of the Chinese Academy of Sciences}
\icmlaffiliation{pcl}{Peng Cheng Laboratory}

\icmlcorrespondingauthor{Changsheng Xu}{csxu@nlpr.ia.ac.cn}

\icmlkeywords{Decoupled Vision System, Multi-modal Large Language Model, Vision-Language Foundation Models}

\vskip 0.3in

]



\printAffiliationsAndNotice{}  

\normalem
\begin{abstract}
In this work, we introduce \textbf{Libra}, a prototype model with a decoupled vision system on a large language model (LLM). The decoupled vision system decouples inner-modal modeling and cross-modal interaction, yielding unique visual information modeling and effective cross-modal comprehension. Libra is trained through discrete auto-regressive modeling on both vision and language inputs. Specifically, we incorporate a routed visual expert with a cross-modal bridge module into a pretrained LLM to route the vision and language flows during attention computing to enable different attention patterns in inner-modal modeling and cross-modal interaction scenarios. Experimental results demonstrate that the dedicated design of Libra achieves a strong MLLM baseline that rivals existing works in the image-to-text scenario with merely 50 million training data, providing a new perspective for future multimodal foundation models. 
Code is available at \url{https://github.com/YifanXu74/Libra}.
\end{abstract}

\section{Introduction}
\label{sec:intro}
The integration of vision and language plays a vital role in machine perception and understanding of the world. Language serves as the basis for cognitive processing, while vision provides essential sensory information. In this context, the field of multimodal large language models (MLLMs)~\cite{yin2023survey} has made remarkable progress, yielding impressive results across various domains, including multimodal conversation~\cite{gemini,gpt4v}, interactive agents~\cite{cogagent}, and even autonomous driving~\cite{cui2024survey}.

A line of recent works~\cite{unified-io,unified-io2,kosmos,kosmos2,ofa} jointly trains multimodal models from scratch, naturally aligning vision and language under the unified structure design and modeling paradigm.
However, these approaches often compromise on unified but not general models due to an information imbalance: general intelligence in the era of foundation models demands a large scale of language knowledge, but unfortunately, visual data falls short in matching the scale of language. For instance, Unified-IO~\cite{unified-io} achieves cross-modal comprehension by making sacrifices in certain language capabilities, especially the wide range of world knowledge and chat ability.

In light of this, another line of works~\cite{flamingo,llava,qwen-vl,emu}, LLM-based approaches, follows a staged training paradigm: first training a large language model (LLM) to acquire a wide range of general knowledge, then integrating visual perception into the pretrained LLM.
This paradigm is reasonable because it can efficiently transfer the general knowledge learned by the language model to the MLLMs.
To this end, on the basis of well-built language systems like LLaMA~\cite{llama2}, building an effective vision system for essential visual sensory and a reasonable cross-modal interaction strategy for cross-modal comprehension upon LLMs becomes a natural idea.

A straightforward approach in most recent LLM-based works is to employ a pretrained vision encoder like CLIP~\cite{clip} as the vision system, integrating its features into a pretrained LLM  to facilitate cross-modal interaction, \emph{e.g.}, through a trainable Q-Former~\cite{blip2} or a simple projection layer~\cite{llava}.
This integration is achieved through an image-captioning loss, where the supervision is only performed on the language part.
However, this pipeline leads to a weak vision system because its visual understanding ability is limited by the pretrained vision encoder.
To address this, several works~\cite{dreamllm}, such as Emu~\cite{emu}, attempt to directly build more sophisticated vision systems on LLMs.
They perform contiguous auto-regressive image modeling, where each input visual feature predicts the input feature of the next position.
Despite impressive image generation results, these works provide limited benefits to downstream tasks because 1) the unified architecture makes coupled vision and language systems, thereby losing unique visual information; 2) the contiguous vision supervision raises an infinite label space that increases the learning difficulty.

In this work, we aim to build a more reasonable vision system upon LLMs.
From a biological perspective~\cite{thiebaut2022emergent}, vision and language systems can exist independently, while vision-language comprehension requires further cross-modal interaction.
This inspires us to consider \textit{what is an ideal vision system on LLMs}.
We believe that the following two aspects are equally important.
1) To retain an extensive and in-depth visual understanding ability, the vision system should be relatively independent from the language model due to the information imbalance.
2) To facilitate cross-modal comprehension, vision systems should be altruistic in aligning the vision and language features.
Based on the above inspiration, we propose to learn a decoupled vision system on LLMs, and build up a new prototype MLLM model \textbf{Libra}.
We found that Libra is a strong MLLM baseline with limited training data (50M in this work vs. 1B in previous works~\cite{li2023multimodal}).
The decoupled vision system of Libra can simultaneously retain unique visual information and support the cross-modal interaction,
which is achieved by the following designs.

\textbf{Routed visual expert.}
The core of the decoupled vision system is a routed visual expert module relied on LLMs, which comprises a simple visual expert and a cross-modal bridge module.
Firstly, the visual expert has its own vision-specific parameters.
It resembles a mixture of experts (MoE)~\cite{moe1,moe2} structure, featuring an additional attention layer and a feed-forward network (FFN) for vision features alongside the existing frozen layers in the LLM for language features.
Secondly, the cross-modal bridge module enables cross-modal interaction, routing the vision and language flows during attention computing to enable different attention patterns in inner-modal modeling and cross-modal interaction scenarios.

\textbf{Discrete auto-regressive modeling.}
The vision system of Libra is learned through a discrete next-token-prediction paradigm on vision inputs, enabling a finite label space for stable learning of the vision system compared to previous contiguous image modeling approaches~\cite{emu,emu2}.
We focus on the image-to-text scenario, where the vision system (routed visual expert) learns unconditional image modeling, and the language system (LLM) learns vision-conditioned language modeling.

\textbf{Hybrid image tokenization.}
A side effect of discrete auto-regressive modeling is the information loss brought by image discretization.
To mitigate this, we propose a hybrid tokenization strategy that combines contiguous visual signals from the vision encoder with discrete modeling using tokenized ids. To leverage the pretrained knowledge of well-established vision encoders like CLIP~\cite{clip}, we construct a CLIP-based image tokenizer using lookup-free quantization (LFQ)~\cite{lfq}. This is the first time that a highly reconstructive image tokenizer can be constructed upon a frozen vision encoder like CLIP, which has not even been investigated in the work of LFQ.

With the dedicated designs, we demonstrate some noteworthy behaviors:
\begin{itemize}
\item We provide a new perspective for the design of MLLMs by modeling a decoupled vision system that decouples inner-modal modeling and cross-modal interaction.
\item The decoupled vision system enhances the attention diversity across layers, reducing the learning redundancy and improving vision-language comprehension.
\item Libra rivals modern MLLMs across more than 15 multimodal benchmarks, despite limited training data.
\end{itemize}

\begin{figure*}[t]
    \vskip 0.2in
    \centering
    \includegraphics[width=0.9\textwidth]{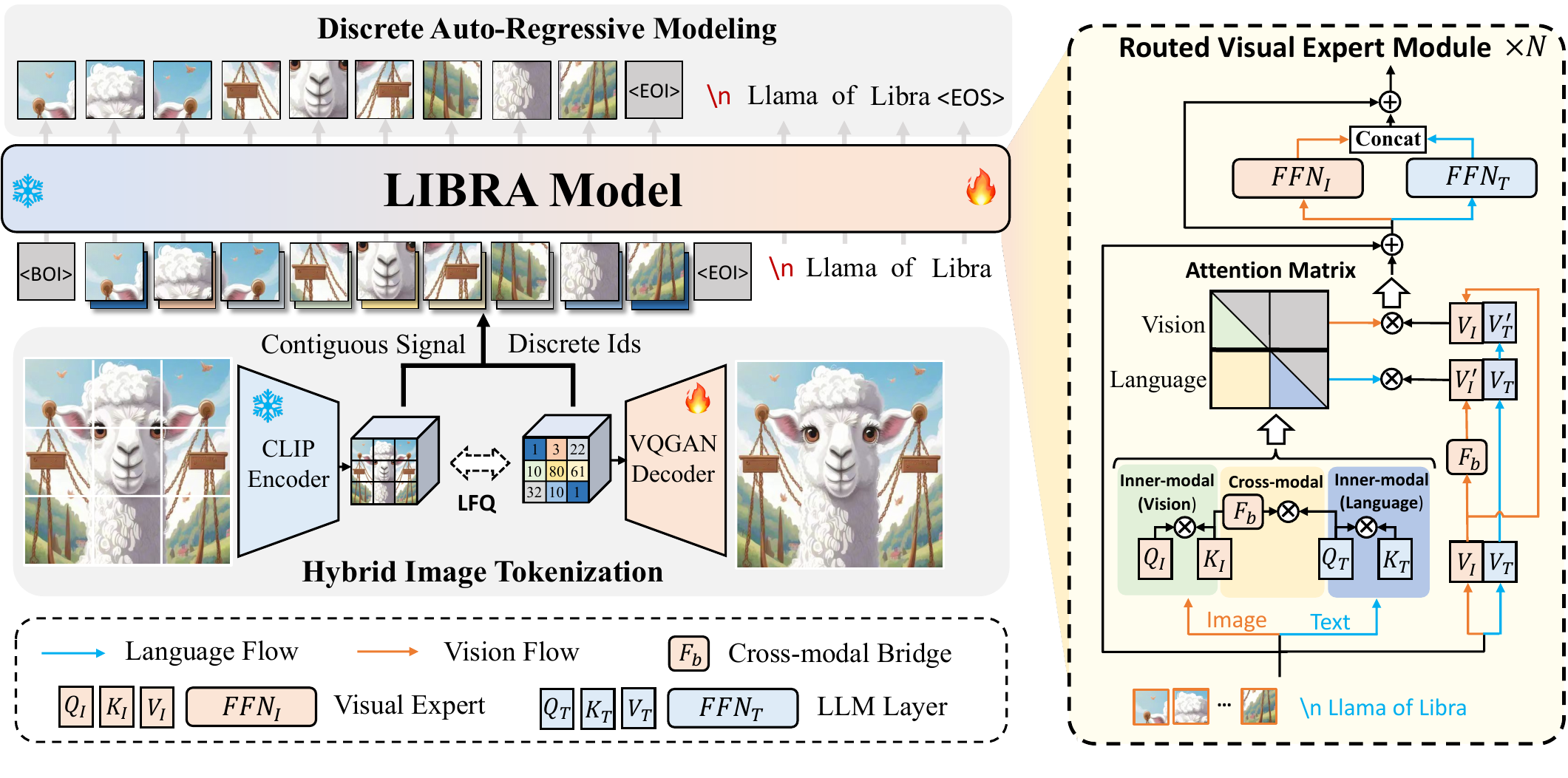}
    \caption{Libra investigates a decoupled vision system on the pretrained LLM. The vision system is built with a routed visual expert design. We train Libra through discrete auto-regressive modeling. The vision inputs consist of a hybrid of contiguous signals from the vision encoder and discrete ``word'' embeddings constructed based on the tokenized ids. $\texttt{<EOS>}$ is the end-of-sequence token. 
    In practice, the discrete ids are used to construct discrete vision embeddings from a codebook learned by auto-regressive image modeling of Libra.
    }
    \label{fig:overview}
\end{figure*}

\section{Related Work}
\label{sec:related_work}
Rapid developments have been witnessed in multimodal large language models (MLLMs)~\cite{yin2023survey} that enable human interaction with both words and visual content. One line of works~\cite{visual_chatgpt,gupta2023visual,shen2023hugginggpt,suris2023vipergpt,yang2023mm} utilizes LLMs as central controllers, integrating them with various functional agents, with language serving as a general interface. This plugin-style framework achieves remarkable success with very low training cost. Another line of works explores directly training MLLMs, including scratch training with unified architectures~\cite{ofa,unified-io,gemini}, integrating pretrained vision encoders with pretrained LLMs through simple projections~\cite{llava,cogvlm,qwen-vl,otter,dreamllm,emu} or cross-attention~\cite{flamingo}.
Several training strategies are proposed to reduce the training burden, including instruction tuning~\cite{xu2022multiinstruct,llava} and parameter-efficient tuning~\cite{lora,dettmers2023qlora,llama-adapter}.

The most related studies to our work are Emu~\cite{emu} and CogVLM~\cite{cogvlm}. Emu performs contiguous auto-regressive image modeling using a CLIP vision encoder and a diffusion~\cite{latent_diffusion} image decoder. 
CogVLM proposes a visual expert module on frozen LLMs to achieve deeper alignment between vision and language. 
Both works demonstrate that the contiguous modeling paradigm does not provide evident benefits for image-to-text vision-language comprehension, despite remarkable text-to-image generation capability.
Instead, we show the importance of stable discrete image modeling with a reasonable cross-modal interaction strategy, which enables an effective vision system on LLMs, ultimately enhancing vision-language comprehension.

\section{Approach}
\label{sec:approach}
\subsection{Architecture}
\label{sec:architecture}
Libra comprises three fundamental components: a unified input tokenizer, a pretrained large language model, and a routed visual expert with a cross-modal bridge module. Fig.\,\ref{fig:overview} illustrates an overview of Libra.
The implementation details are introduced as follows.

\textbf{Unified Input Tokenizer.}
Libra unifies both vision and language modeling into a \textit{discrete} next-token-prediction paradigm. Given an input sequence with both image and corresponding language parts, we separately tokenize the image and the language parts into discrete tokens through a VQGAN~\cite{vqgan} and a SentencePiece~\cite{sentencepiece} tokenizer. We respectively prefix and suffix the image sequence with a $\langle \textrm{BOI} \rangle$ (beginning of image) token and a $\langle \textrm{EOI} \rangle$ (end of image) token. 
We use a newline token ``\verb|\n|'' to separate images and texts.
All token embeddings except the separation newline token ``\verb|\n|'' are updated through a cross-entropy classification loss.
Meanwhile, integrating images into completely discrete tokens results in severe information loss, as verified in Sec.\,\ref{sec:discuss}. Therefore, we propose a hybrid image tokenization process (Sec.\,\ref{sec:tokenization}) to enable both stable discrete sequential modeling and contiguous visual comprehension.

\textbf{Pretrained Large Language Model.}
Libra's model design is compatible with any off-the-shell GPT-style pretrained LLMs. We adopt the commonly used LLaMA2-7B-Chat~\cite{llama2} for further training. We freeze the LLM during pretraining and unfreeze it during instruction tuning (see Sec.\,\ref{sec:training}).

\textbf{Routed Visual Expert.}
As described in Sec.\,\ref{sec:intro}, the LLM-based approaches are built upon a wide range of general knowledge brought by the LLM. A decoupled vision system can preserve unique visual information without distorting the inherent knowledge within the LLM. Therefore, we propose a routed visual expert for vision-specific encoding and decoupled cross-modal interaction.
We add the routed visual expert to each layer of the LLM and freeze the LLM during pretraining to preserve its language knowledge. 
 
The routed visual expert features: 1) additional attention and FFN layers for vision features alongside the original LLM layers for language features, and 2) a cross-modal bridge for cross-modal interaction. Formally, given the input hidden states $X \in \mathbb{R}^{B \times H \times (L_{I} + L_{T}) \times D}$ with the image part $X_{I}$ of length $L_{I}$ and the text part $X_{T}$ of length $L_{T}$, where $B$ is the batch size, $H$ is the number of attention heads, and $D$ is the hidden size. The attention is computed as:
\begin{equation}
\begin{split}
X_{I}^{a}, X_{T}^{a} &= \operatorname{Attn}\left(X\right) = \\
&\operatorname{softmax}\left(\frac{\operatorname{Tril}\left(Q \cdot F_b(K)^T\right)}{\sqrt{D}}\right) F_{b}(V),\\
Q&=\operatorname{concat}\left(X_I W_I^Q, X_T W_T^Q\right),\\
K&=\operatorname{concat}\left(X_I W_I^K, X_T W_T^K\right),\\
V&=\operatorname{concat}\left(X_I W_I^V, X_T W_T^V\right),
\end{split}
\label{eqn:attn_expert}
\end{equation}
where $X_{I}^{a}, X_{T}^{a}$ is the attention outputs of the vision and language parts, respectively. $F_b$ refers to the cross-modal bridge module introduced in the next part, $W_{I}^{*}$, $W_{T}^{*}$ are the QKV matrices of the visual expert and original language model, and $\operatorname{Tril}(\cdot)$ denotes the causal lower-triangular mask. 
For parameter-efficiency, we represent each visual expert matrix $W_{I}^{*} \in \mathbb{R}^{D \times D^{\prime}}$ as the product of two low-rank matrices, namely: $W_{I}^{*}=A_{I}^{*} \cdot B_{I}^{*}$, where $A_{I}^{*} \in \mathbb{R}^{D \times D/4}$ and $B_{I}^{*} \in \mathbb{R}^{D/4 \times D^{\prime}}$.
Similarly, the visual expert in feed-forward network (FFN) layers performs as:
\begin{equation}
\operatorname{FFN}(X)=\operatorname{concat}\left(\operatorname{FFN}_I\left(X_I\right), \operatorname{FFN}_T\left(X_T\right)\right),
\label{eqn:ffn_expert}
\end{equation}
where $\operatorname{FFN}_I$, $\operatorname{FFN}_T$ are the FFNs of the visual expert and the original language model.

It is worth noting that the visual expert design here is similar to the one proposed in CogVLM~\cite{cogvlm}. The differences lie in various aspects. 
1) \textit{Approach}: we further introduce a cross-modal bridge module to decouple inner-modal modeling and cross-modal interaction. 
2) \textit{Goal}: Libra uses the visual expert design as one reasonable path to achieve a decoupled vision system upon frozen LLMs, while CogVLM only uses it for better vision-language alignment.
3) \textit{Insight}: we demonstrate that effective image modeling of the vision system significantly enhances vision-language comprehension under the visual expert design (see Sec.\,\ref{sec:discuss}), in contrast to the findings of CogVLM described in Sec.\,\ref{sec:related_work}.

\textbf{Cross-modal Bridge.}
In addition to the modality-specific modeling brought by the visual expert and the LLM, we found that a decoupled cross-modal interaction strategy plays a vital role in cross-modal comprehension. We observed that image modeling fails with a simple visual expert, indicating an invalid vision system (see Fig.\,\ref{fig:img_modeling}(a)). This is because a simple visual expert  does not really build a decoupled vision system.
In image-to-text scenarios, language predictions are based on the image condition. The frozen LLM places the entire learning burden of cross-modal interaction on the visual expert module. Consequently, the vision modeling of the visual expert tends to align with language, resulting in an inability to learn meaningful visual representations.

Therefore, we design a cross-modal bridge module to decouple the inner-modal modeling and cross-modal interaction. The bridge adds an additional learnable projection upon the keys and values when computing cross-modal attention.
Formally, given the input hidden states $X = [X_{I}, X_{T}]$, the attention keys of the vision part are computed as:
\begin{equation}
\begin{gathered}
    F_b\left(K_{I} \mid Q_{*}, X_{I}\right) = \begin{cases}K_{I} & ,\text{ if inner-modal}, \\ K_I^{\prime} & ,\text{ if cross-modal},\end{cases} \\
    K_I^{\prime} = K_I+X_I W_{I}^{K\prime},
\end{gathered}
\label{eqn:bridge}
\end{equation}
where $K_I = X_I W_{I}^{K}$ in Eqn.\,(\ref{eqn:attn_expert}), $W_{I}^{K\prime}$ is the learnable transformation projection of the bridge module. Eqn.\,(\ref{eqn:bridge}) denotes that: we transform the original keys to new values if $Q_{*}$ and $K_{I}$ are from different modalities (cross-modal); if $Q_{*}$ and $K_{I}$ are from the same modality (inner-modal), we keep the original keys.
Note that the condition suffix of $F_b(\cdot)$ in Eqn.\,(\ref{eqn:bridge}) is omitted in the other parts of the paper for concision. Similarly, we can get the attention keys $F_b(K_{T})$ of the text part, which is unused in the image-to-text scenario. Finally, the attention matrix can be computed as:
\begin{equation}
Q \cdot F_b(K) = \left[\begin{array}{ll}
Q_I K_I^{\top} & Q_I K_T^{\prime \top} \\
Q_T K_I^{\prime\top} & Q_T K_T^{\top}
\end{array}\right].
\end{equation}
Similarly, the computing of the bridge module on attention values can be illustrated through Fig.\,\ref{fig:overview}, where we transform and keep the original values under cross-modal and inner-modal scenarios, respectively. Formally,
\begin{equation}
\begin{split}
    X_{I}^{a} &= \sigma \left(\left[\begin{array}{ll}Q_I K_I^{\top} & Q_I K_T^{\prime \top} \end{array}\right] \right) \cdot \left[\begin{array}{l}V_{I} \\ V_{T}^{\prime} \end{array}\right], \\
    X_{T}^{a} &= \sigma \left(\left[\begin{array}{ll}Q_T K_I^{\prime \top} & Q_T K_T^{\top} \end{array}\right] \right) \cdot \left[\begin{array}{l}V_{I}^{\prime} \\ V_{T} \end{array}\right], \\
    V_{I}^{\prime} &= V_I + X_I W_I^{V \prime}, \\
    V_{T}^{\prime} &= V_T + X_T W_T^{V \prime},
\end{split}
\end{equation}

where $V_I=X_I W_I^V$ and $V_T=X_T W_T^V$ in Eqn.\,(\ref{eqn:attn_expert}). $W_{I}^{V\prime}$ and $W_{T}^{V\prime}$ are the learnable transformation projections of the bridge module. $\sigma$ denotes a softmax function. We omit the normalization factor and the causal mask for concision. In practice, $V_{T}^{\prime}$ takes no effect during pretraining due to the causality of auto-regression, as the data are always formulated as
\texttt{<Image>}\verb|\n| \texttt{<Text>}, with visuals preceding the text.

Last but not least, the transformation brought by the bridge module should not be too large, in order to leverage the learned knowledge in the original keys and values. Thus, we apply a low-rank strategy to the design of the transformation projection $W_{I}^{*\prime}, W_{T}^{*\prime} \in \mathbb{R}^{D \times D^{\prime}}$. Take the vision part as example: $W_{I}^{*\prime} = A_I^{*\prime} \cdot B_I^{*\prime}$, where $A_I^{*\prime} \in \mathbb{R}^{D \times 8}$ and $B_I^{*\prime} \in \mathbb{R}^{8 \times D^{\prime}}$.

\subsection{Hybrid Image Tokenization}
\label{sec:tokenization}
The unified discrete next-token-prediction paradigm raises two obstacles for effective vision-language comprehension. 1) The discretization process of VQGAN can cause severe visual information loss, leading to low perception on visual details. 2) Naive discrete sequential modeling hardly benefits from the pretrained knowledge of the vision encoder, since the model receives newly-constructed embeddings based on the input ids instead of the features of the vision encoder. 
To this end, as shown in Fig.\,\ref{fig:overview}, we propose a hybrid image tokenization process from two aspects: contiguous visual signals and pretrained visual knowledge. 

\begin{figure}[t]
    \vskip 0.1in
    \centering
    \includegraphics[width=0.45\textwidth]{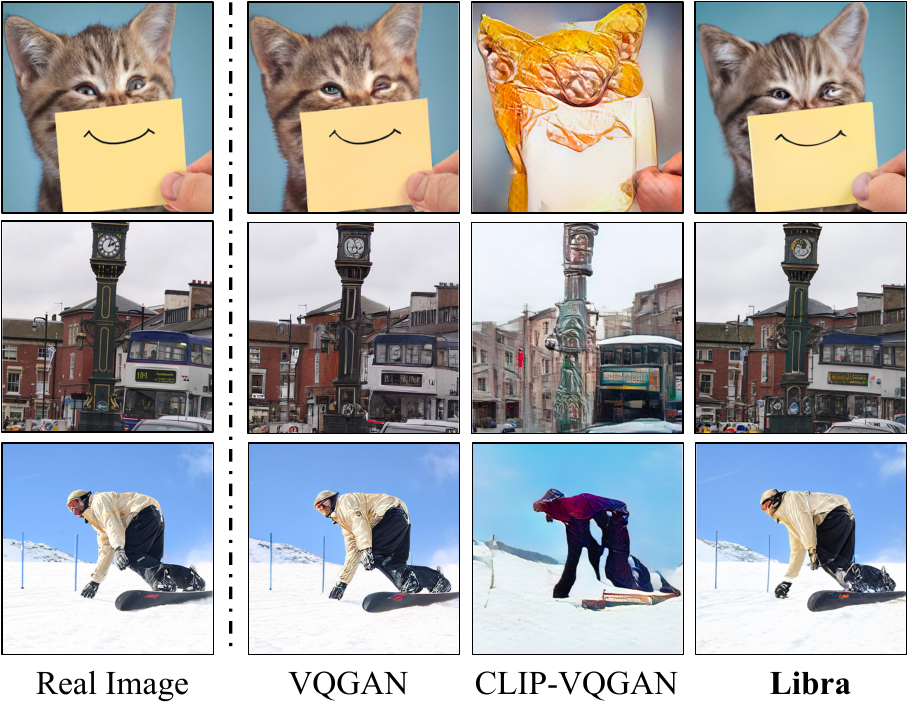}
    \caption{Image reconstruction results. Directly replacing the image encoder of VQGAN with CLIP distorts the visual information. Libra largely alleviates this problem via lookup-free quantization.}
    \label{fig:vis_tokenize}
\end{figure}

\textbf{Contiguous Visual Signal vs. Discrete Modeling.}
We leverage a hybrid tokenization strategy with a combination of contiguous visual signals from the vision encoder and discrete modeling using tokenized ids. Specifically, as illustrated in Fig.\,\ref{fig:overview}, given an input image, we first feed it into the vision encoder and obtain the output features as contiguous visual signals. Then, a quantization/discretization process is performed to tokenize the contiguous visual signals into discrete token ids based on a vision codebook. The token ids are used to construct vision ``word'' embeddings similar to the ones in LLMs. Finally, we concatenate the contiguous visual signals and the discrete vision embeddings in the channel dimension as the final vision inputs of the Libra model. Sec.\,\ref{sec:tokenizer_appendix} provides more details of the tokenization process.
Meanwhile, a discrete auto-regressive image modeling is performed on the output features of Libra, \emph{i.e.}, each vision input is used to predict the token id of the next position.
This simple design enables both contiguous visual comprehension and stable discrete sequential modeling.

\textbf{Pretrained Visual Knowledge.}
To leverage the pretrained knowledge in existing well-established vision encoders like CLIP~\cite{clip}, we replace the vision encoder of VQGAN with a \textit{frozen} CLIP-ViT-L-336px. However, training a CLIP-based VQGAN is non-trivial. The features in CLIP are highly semantic with less low-level visual information. Directly emulating such features in the quantization process of the original VQGAN results in poor reconstruction performance, as demonstrated in Fig.\,\ref{fig:vis_tokenize} (CLIP-VQGAN). Instead, we find that the lookup-free quantization (LFQ)~\cite{lfq}, which does not need to emulate the input features, can largely address this problem. To this end, we use a CLIP-based VQGAN with LFQ as Libra's image tokenizer (Libra in Fig.\,\ref{fig:vis_tokenize}). This is the first time that a highly reconstructive image tokenizer can be constructed based on a frozen vision encoder like CLIP, which has not even been investigated in the work of LFQ.

We train our image tokenizer using 10M images collected by \cite{sam}. The vision vocabulary size is enlarged to $2^{18}$ thanks to LFQ. For computational efficiency, we predict in two concatenated codebooks, each of size $2^{9}$. More details can be found in Sec.\,\ref{sec:tokenizer_appendix}.

\subsection{Training}
\label{sec:training}
\textbf{Pretraining.}
Libra is pretrained under unified sequential modeling, where a next-token-prediction objective is performed on all input tokens, as:
\begin{equation}
p(X)=\prod_{\ell=1}^L p\left(X_{\ell} \mid X_{<\ell}\right),
\end{equation}
where $X=[X_I,X_T]$ is the input multimodal sequence and $L$ is the sequence length. In practice, the objective is computed through a discrete cross-entropy classification loss. 
We use image-text pairs for training.
We freeze the LLM and only update the routed visual expert during pretraining.

\textbf{Multimodal Instruction Tuning.}
Language instruction tuning has helped LLMs to align with user intentions~\cite{rlhf,self-instruct} and generalize to unseen tasks~\cite{wei2021finetuned,chung2022scaling}. Similarly, we apply multimodal instruction tuning on the pretrained Libra model. 
We train the whole model during tuning.
All instruction-tuning data are arranged based on this template:
\begin{equation}
    \begin{split}
        &\prompt{\text{<System Message>}}\\
        &\prompt{\text{[USER]: <Image> <Instruction>}}\\
        &\prompt{\text{[ASSISTANT]: <Answer>}}\\
    \end{split}
    \label{eqn:templete}
\end{equation}
where only \texttt{<Answer>} is accounted for computing loss, as:
\begin{equation}
p(X_{a} \mid X_{v}, X_{instruct})=\prod_{\ell=1}^L p\left(x_{\ell} \mid X_{v}, X_{instruct}, x_{<\ell}\right),
\end{equation}
where $X_{a} = \{x_{\ell}\}_{\ell=1}^{L}$, $X_v$, $X_{instruct}$ are the answers, images, and instructions.

\begin{table*}[t]
\caption{Performance comparison on visual question answering (VQA) and image captioning. 
Specialists perform dataset-specific finetuning, while generalists commonly perform zero-shot evaluation.
The pretraining data sizes are reported.  $^{*}$The training images of the datasets are observed during training. $^{\dagger}$Includes in-house data that is not publicly accessible.}
\label{tab:vqa}
\vskip 0.1in
\begin{center}
\begin{small}
\setlength{\tabcolsep}{1mm}{
\begin{tabular}{l|ll|lllll|lll}
\toprule
            &                              &                            & \multicolumn{5}{c|}{General VQA}                                                                                                                                          & \multicolumn{3}{c}{Image Caption}                                                                                 \\
Method      & \#Params & \#Data & VQAv2 & OKVQA & GQA & VizWiz & SQA & NoCaps & Flickr & COCO \\ \midrule
\multicolumn{11}{c}{\textit{Specialists}}                                                                                                                                                                                                                                                                                                                                       \\ \midrule
BEiT-3~\cite{beit3}      &        1.9B               &          50M                  &              84.0          &           -             &       -              &            -             &            -           &                 -          &              -            &            147.6              \\
PaLI-X~\cite{palix}             &          55B                &   -                      &       86.1              &          66.1              &           -            &         70.9                 &          -               &                  124.3    &           -           &           149.2               \\
OFA~\cite{ofa}      &        930M                      &              60M              &               82.0         &            -             &           -             &                     -       &            -             &                      -     &           -              &           154.9          \\
CogVLM~\cite{cogvlm}      &           17B             &     1.5B$^{\dagger}$                       &               84.7         &            64.7             &           65.2             &                     75.8       &            92.7             &                      126.4     &           94.9              &           144.9          \\

\midrule

\multicolumn{11}{c}{\textit{Generalists}}                                                                                                                                                                                                                                                                                                                                       \\ \midrule
BLIP-2~\cite{blip2}    &         12.1B                     &           129M               &          65.0               &           45.9             &           44.7          &                -            &           -               &                  121.6        &             74.9             &           \textbf{144.5}$^{*}$           \\
Flamingo~\cite{flamingo}    &        80B                &          2.1B$^{\dagger}$              &            56.3           &          50.6            &             -               &                 31.6      &               -            &    -                     &        67.2              &             84.3             \\
$\text{Unified-IO}_{\text{XL}}$~\cite{unified-io}    &       2.9B                 &                    -       &          77.9$^{*}$             &           54.0$^{*}$           &       -                     &          \uline{57.4}$^{*}$             &        -                  &          100.0              &           -            &        122.3$^{*}$                  \\
PaLM-E~\cite{palme}    &        12B                  &       70M$^{\dagger}$         &   
      76.2$^{*}$                  &                55.5$^{*}$         &             -          &                 -          &                   -       &       -                    &              -           &             135.0$^{*}$               \\
InstructBLIP~\cite{instructblip}    &            14.2B             &       129M                    &          -               &               -       &            49.5               &            33.4           &      63.1                   &               \uline{121.9}          &          \uline{82.8}            &       104.2$^{*}$                    \\
Emu~\cite{emu}    &            14B              &           4B                &            40.0$^{*}$            &         34.7            &               -            &          35.4              &                  -         &          -               &         -               &            117.7$^{*}$              \\
Qwen-VL~\cite{qwen-vl}    &              9.6B          &               1.4B$^{\dagger}$         &           \uline{78.2}$^{*}$            &       \uline{56.6}              &           57.5$^{*}$                &          38.9              &              68.2            &                 120.2        &         81.0             &          -                \\
Shikra~\cite{shikra}   &        13.3B               &       600K                    &          77.4$^{*}$              &         53.8             &        -                   &         -               &               -            &          -                &           73.9$^{*}$            &          117.5$^{*}$              \\
IDEFICS~\cite{idefics}    &         80B                &     353M                    &          60.0              &         -              &         45.2                 &         36.0              &               -            &      -                   &           -          &        -                   \\
LLaVA1.5~\cite{llava1.5}    &      13.4B                   &            558K           &          \textbf{80.0}$^{*}$             &        -               &          \uline{63.3}$^{*}$                 &            53.6            &              \uline{71.6}           &      -                     &        -                 &             129.8$^{*}$               \\
\midrule
\textbf{Libra (ours)} &         11.3B                 &       50M                 &      77.3$^{*}$                  &           \textbf{59.7}              &            \textbf{63.8}$^{*}$                  &         \textbf{59.5}             &               \textbf{73.5}             &                   \textbf{123.8}         &        \textbf{86.6}               &       \uline{135.2$^{*}$}                      \\
\midrule
\end{tabular}
}
\end{small}
\end{center}
\vskip -0.1in
\end{table*}

\textbf{Data.}
In this work, we only build Libra as a prototype model, thereby using much less pretraining data than most of previous works. 
For pretraining, we use 50M image-text pairs randomly sampled from COYO-700M~\cite{coyo700m} and CC12M~\cite{cc12m}. We use additional 500K image-text pairs from COCO~\cite{cococap} training split to standardize the caption outputs.
For instruction tuning, we leverage the 665K high-quality supervised  data from LLaVA-Instruct~\cite{llava}.
More training details can be found in Sec\,\ref{sec:training_appendix}.

\section{Experiments}

\subsection{Implementation}
Libra consists of 11.3 billion parameters, with 7 billion from the LLM, 4 billion from the routed visual expert, and 0.3 billion from the CLIP vision encoder.
We conduct comprehensive evaluation on various tasks, including visual question answering (VQA), image captioning, and MLLM-oriented multimodal benchmarks. We refer to Sec.\,\ref{sec:benchmark_appendix} for more details on the evaluation benchmarks and metrics. All evaluations are performed based on greedy search for replication. 

\begin{table*}[t]
\caption{The zero-shot evaluation on MLLM-oriented multimodal benchmarks.}
\label{tab:benchmark}
\vskip 0.1in
\begin{center}
\begin{small}
\setlength{\tabcolsep}{1.4mm}{
\begin{tabular}{l|l|l|llllllll}
\toprule
Method      & LLM & \#Data & POPE & MME & $\text{MME}^{C}$ & MMB & $\text{MMB}^{CN}$ & SEED & MM-Vet & MMVP \\
\midrule
BLIP-2~\cite{blip2}    &  $\text{FlanT5}$-11B   &   129M     &  85.3    &   1293.8  &    290.0    &  -   &   -   &   46.4   &    22.4    &   -   \\
InstructBLIP~\cite{instructblip}      &  $\text{FlanT5}$-11B   &    129M    &   78.9   &  1212.8   &    \uline{291.8}    &   -  &   -   &   -   &    -    &   16.7   \\
Shikra~\cite{shikra}      &   Vicuna-13B  &    600K    &   -   &   -  &   -     &   58.8  &   -   &  -    &   -     &   -   \\
MiniGPT-4~\cite{minigpt4}      &   Vicuna-13B  &   3.5K     &  -    &  581.6   &   144.2     &   23.0  &   -   &   42.8   &    22.1    &   12.7   \\
Otter~\cite{otter}      &  LLaMA-7B   &    2.1B    &   -   &  1292.2   &    -    &   48.3  &   -   &   32.9   &    24.6    &   -   \\
IDEFICS~\cite{idefics}      &   LLaMA-65B  &   353M     &   -   &  -   &   54.5     &   38.1  &   -   &   -   &    -    &  -    \\
mPLUG-Owl~\cite{mplug-owl}      &  LLaMA-7B   &   1.2B     &   -   &  967.3   &    276.0    &  46.6   &   -   &   34.0   &    -    &   -   \\
Qwen-VL~\cite{qwen-vl}      &   Qwen-7B  &    1.4B    &   -   &  1487.5   &    \textbf{360.7}    &   60.6  &   56.7   &   58.2   &   -     &    -  \\
LLava1.5~\cite{llava1.5}      &   Vicuna-7B  &   558K     &  85.9    &   \textbf{1510.7}  &   274.3     &   \uline{64.3}  &  \uline{58.3}    &  \uline{58.6}    &    \uline{30.5}    &   \uline{24.7}   \\

\midrule
\textbf{Libra (ours)} &  LLaMA2-7B   &   50M     &    \textbf{88.2}  &   \uline{1494.7}  &   281.1     &  \textbf{65.2}   &   \textbf{58.8}   &    \textbf{62.7}  &     \textbf{31.8}   &  \textbf{30.0} \\
\midrule
\multicolumn{11}{c}{\textit{\g{Commercial Chatbots}}} \\ 
\midrule
\g{GPT-4V~\cite{gpt4v}} &  \g{-}   &   \g{-}     &  \g{-}    &   \g{1409.4}  &   \g{517.1}     &   \g{77.0}  &   \g{74.4}   &   \g{71.6}   &   \g{-}     &  \g{38.7} \\
\g{Gemini-Pro~\cite{gemini}} &  \g{-}   &   \g{-}     &   \g{-}   &  \g{1496.5}   &    \g{436.7}    &   \g{73.6}  &  \g{74.3}    &   \g{70.7}   &     \g{-}   & \g{40.7}  \\
\g{Bard~\cite{bard}} &  \g{-}   &   \g{-}    &   \g{-}   &  \g{-}   &   \g{-}     &  \g{-}   &   \g{-}   &   \g{-}   &    \g{-}    &  \g{19.0} \\
\bottomrule
\end{tabular}
}
\end{small}
\end{center}
\end{table*}
\begin{figure*}[t]
    \centering
    \includegraphics[width=0.9\textwidth]{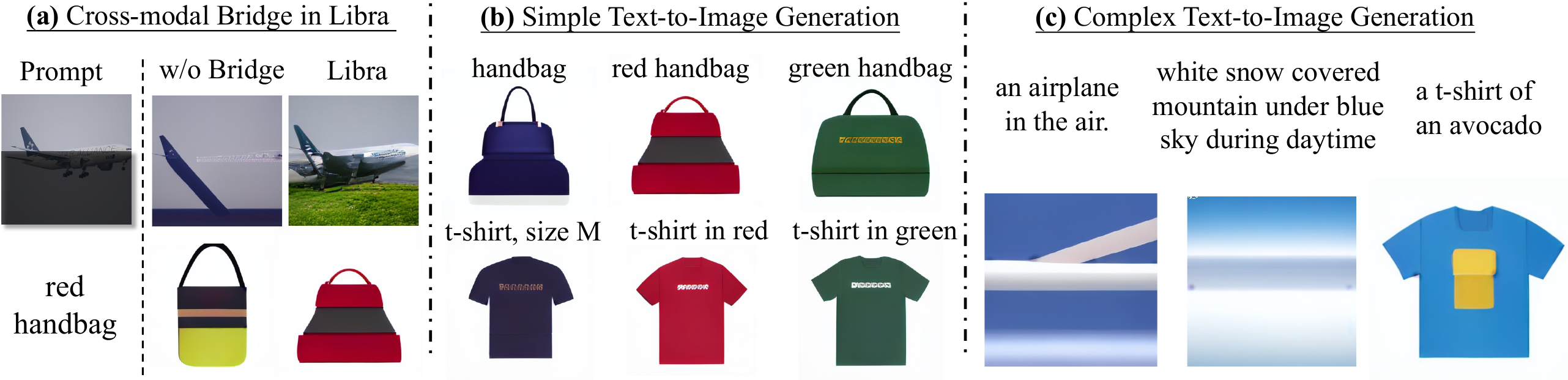}
    \caption{Results of visual sequential modeling.}
    \label{fig:img_modeling}
    \vskip -0.1in
\end{figure*}

\subsection{Vision-Language Comprehension}
\label{sec:vl_comprehension}
\textbf{Visual Question Answering and Image Captioning.}
We evaluate Libra on a wide range of academic benchmarks, including 5 popular general VQA benchmarks and 3 image captioning benchmarks. Tab.\,\ref{tab:vqa} shows the results. 
Libra exhibits strong generalization capabilities in zero-shot captioning and question answering tasks, surpassing previous generalist models with more parameters or larger pretraining data sizes, \emph{e.g.}, it achieves a notable improvement of +20.6\% on the VizWiz dataset compared to Qwen-VL, despite using only 4\% of the pretraining data. Moreover, when using the same instruction tuning data, Libra outperforms LLaVA1.5 on zero-shot tasks, indicating the effectiveness of the vision system in Libra.

\textbf{MLLM-oriented Multimodal Benchmarks.}
Recent studies~\cite{mme,mmb} found that traditional academic benchmarks often fall short in providing a comprehensive ability assessment. To fully evaluate the generality of MLLMs, research communities have introduced a series of benchmarks. We evaluate Libra on 8 MLLM-oriented multimodal benchmarks in Tab.\,\ref{tab:benchmark}. We highlight the best and second-best results in bold and underlined, respectively. The results confirm that Libra rivals existing modern MLLMs.

\subsection{Visual Sequential Modeling}
Despite promising results in vision-language comprehension tasks in Sec.\,\ref{sec:vl_comprehension}, the metrics cannot directly reflect the effectiveness of the vision system. A vision system with good visual representation should at least learn the basic image distribution. Therefore, we examine the vision system of Libra from the perspectives of image completion and text-to-image generation. 
We disable the contiguous visual signal (see Sec.\,\ref{sec:tokenization}) by replacing it with zero values to enable image generation. For text-to-image generation, we further finetune Libra with additional 10 million text-image pairs (7B tokens in total) from the pretraining data. 
Note that our aim here is to validate the effectiveness rather than striving for state-of-the-art performance.

\textbf{Cross-modal Interaction with Cross-modal Bridge.}
Fig.\,\ref{fig:img_modeling}(a) shows the image completion and text-to-image generation results of Libra and its variant without the cross-modal bridge module.
The results show that the variant without the cross-modal bridge module learns a coupled and weak vision system, which: 1) only learns repetitive patterns in image completion, and 2) hardly follows the language instruction during text-to-image generation. A reasonable cross-modal interaction strategy brought by the cross-modal bridge largely boosts effective vision system learning on an LLM. Tab.\,\ref{tab:ablation}(e) also quantitatively shows the effectiveness of the cross-modal bridge module.

\textbf{Naive Image Generation.}
The text-to-image generation results in Fig.\,\ref{fig:img_modeling}(b)(c) indicate that Libra can learn basic structures and concepts (\emph{e.g.}, colors). Fig.\,\ref{fig:img_modeling}(c) shows that Libra fails under complex text-to-image generation. This might be due to the limited training data (training tokens: 7B in Libra vs. 400B in DALL-E~\cite{dalle}). Despite naive image generation performance, the results sufficiently prove that Libra learns the basic image distribution.

\begin{table*}[t]
\caption{Ablation results on VQA and MLLM benchmarks. $^{*}$The training images of the datasets are observed during training.}
\label{tab:ablation}
\begin{center}
\begin{small}
\resizebox{\linewidth}{!}{
\setlength{\tabcolsep}{1.5mm}{
\begin{tabular}{lccc >{\centering\arraybackslash}p{0.1cm} c|c|ccc|ccc}
\toprule
Ablated                       &     & Ablated   & Libra           & \multirow{2}{*}{$\rightarrow$} & Changed  &  \multirow{2}{*}{\#Params}   & \multicolumn{3}{c|}{General VQA} & \multicolumn{3}{c}{MLLM Benchmark} \\
Setting                       &     & Details   & Original Value &                   & Value     &  & VQAv2$^{*}$  & GQA$^{*}$  & VizWiz  & MME      & POPE     & SEED     \\
\midrule
\multicolumn{6}{c|}{\textbf{Libra model}}     &  11.3B    &   77.3   &     63.8    &      59.5    &     1494.7    &      88.2   &  62.7 \\ 
\midrule
\multirow{3}{*}{Training}     & \textbf{(a)} & Paradigm      & Unified        &                   & Language   &   11.3B    &  77.0     &    60.8  &    52.9     &      1465.5    &     86.0    &     58.8     \\
                              & \textbf{(b)} & Supervision     & Discrete       &                   & Contiguous &   11.3B    &    77.2    &   60.5   &    53.3     &    1473.4      &      85.2   &     58.2     \\
                              & \textbf{(c)} & \#Data    & 50M             &                   & 3M      &    11.3B     &   67.8     &    54.9  &     48.8    &    1189.8      &   82.1
      &   51.8       \\
\midrule
\multirow{4}{*}{Architecture} & \textbf{(d)} & Expert       & \Checkmark                &                   & \XSolidBrush    &     7.5B    &    77.0    &  61.4    &     50.8    &     1450.2     &     85.9    &     58.4     \\
                              & \textbf{(e)} & Bridge    & \Checkmark              &                   & \XSolidBrush      &     11.2B        &   76.3     &   61.4   &    53.6     &    1458.4      &    86.0     &     59.6     \\
                              & \textbf{(f)}& Input     & Hybrid     &                   & Discrete   &     11.3B        &   65.0     &  53.1    &    38.6     &    1127.8      &     80.3    &   48.5       \\  
                              & \textbf{(g)}& Vision Encoder     & CLIP     &                   & Scratch   &     11.3B        &   67.2     &  55.7    &   40.6     &    1148.4      &     80.7    &   50.8       \\ 
\bottomrule
\end{tabular}
}
}
\end{small}
\end{center}
\end{table*}
\begin{figure*}[t]
    \centering
    \includegraphics[width=0.9\textwidth]{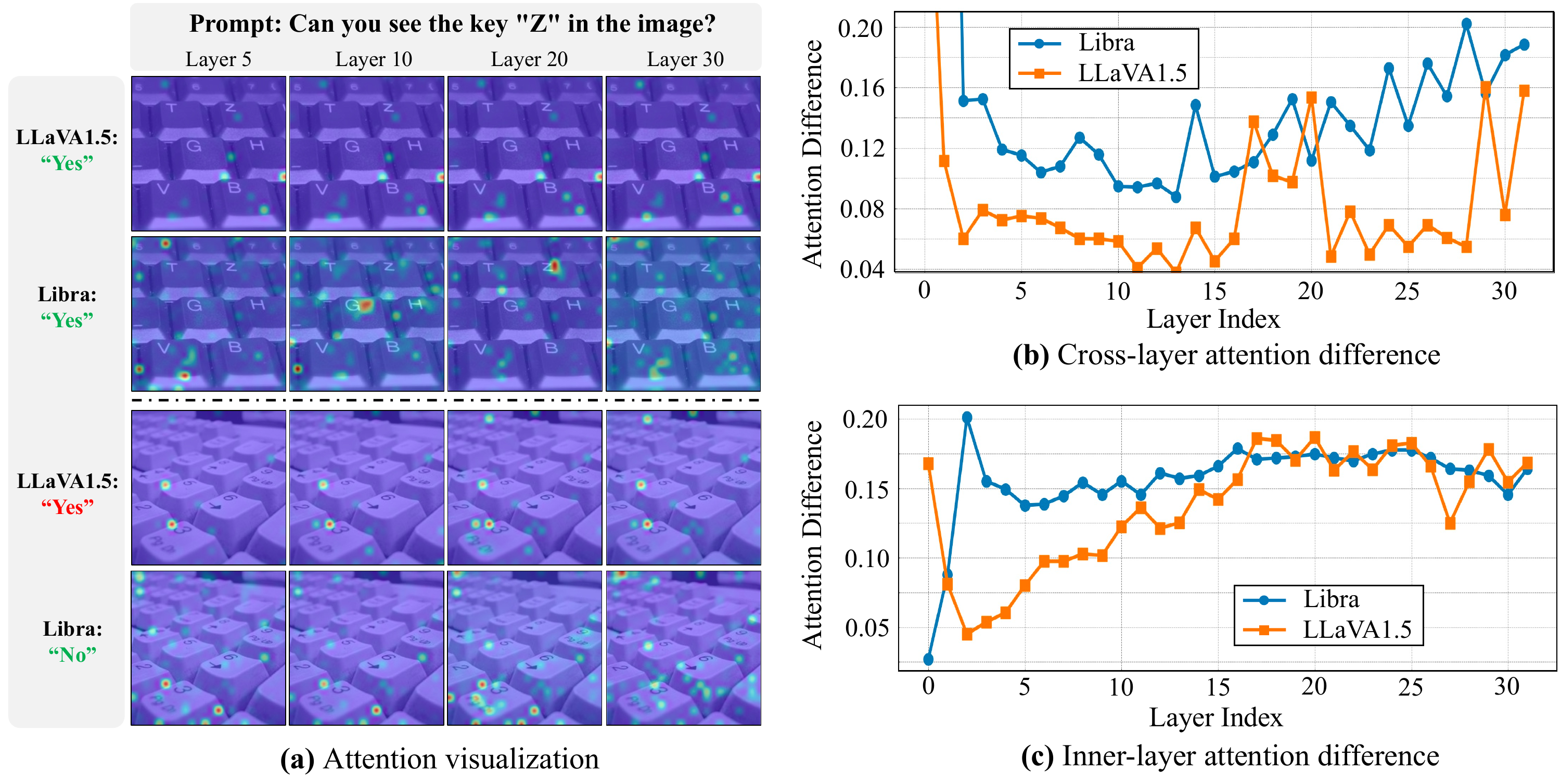}
    \caption{Attention patterns across layers.
    (a) Attention activation of single-word answers on images. (b) Cross-layer attention difference: the difference between each layer's attention score (averaged across all heads) and the mean value of all layers, averaged along the spatial dimension. (c) Inner-layer attention difference: the difference between each head's attention score and the mean value of all heads in each layer, averaged along the spatial dimension. The implementation details can be found in Sec.\,\ref{sec:attn_diff}.}
    \label{fig:vis_attn}
    \vskip -0.1in
\end{figure*}

\subsection{Discussion}
\label{sec:discuss}

\textbf{Impact of a Decoupled Vision System.}
We found that a decoupled vision system impacts the following aspects: 

(1) \textit{Attention diversity}.
We analyze Libra's attention patterns in Fig. \ref{fig:vis_attn}. In Fig. \ref{fig:vis_attn}(a), LLaVA1.5~\cite{llava1.5} shows a consistent attention pattern across layers in VQA, while Libra exhibits diverse attention patterns across layers. The implementation details can be found in Sec.\,\ref{sec:attn_diff}. Quantitative results in Fig. \ref{fig:vis_attn}(b)(c), averaged 100 VQA samples, reveal interesting findings: 1) In Fig. \ref{fig:vis_attn}(b), LLaVA1.5 demonstrates distinct attention patterns only in few middle and deep layers, while Libra shows diverse attention patterns across all layers. 2) In Fig. \ref{fig:vis_attn}(c), LLaVA1.5 exhibits low inner-layer attention differences in shallow layers, whereas Libra diversifies attention patterns across all layers. These results suggest that the decoupled vision system has lower learning redundancy in both cross-layer and inner-layer aspects, as observed through diverse attention patterns.

(2) \textit{Learning bias}.
The MMVP~\cite{mmvp} benchmark in Tab.\,\ref{tab:benchmark} is designed to detect the perception bias in CLIP-based MLLMs, where most MLLMs perform even lower than random guess (25\%). Libra achieves remarkable performance on this benchmark, clearly surpassing previous MLLMs (\emph{e.g.}, +5.3\% over LLaVA1.5). This indicates that a decoupled vision system preserves unique visual information through discrete auto-regressive image modeling. This resembles a regularization effect in previous self-supervised pretraining approaches~\cite{mae,dino}, alleviating the learning bias in MLLMs.

(3) \textit{General performance}.
To verify the impact of image modeling on the vision system, we only supervise the language part of Libra, as shown in Tab.\,\ref{tab:ablation}(a). The results show that Libra has an obvious performance degradation without the discrete image modeling. 
This might be because image modeling encourages unique visual information learning, enabling meaningful visual representation.

\textbf{Discrete Modeling vs. Contiguous Modeling.}
Previous studies~\cite{cogvlm,emu} show that a contiguous image modeling paradigm in pretraining makes limited benefits to downstream tasks, where each visual feature predicts the CLIP feature of the next position for visual self-supervision. In contrast, Libra's discrete image modeling effectively addresses these issues. To validate this, we convert Libra to contiguous image modeling in Tab.\,\ref{tab:ablation}(b). We observed that contiguous image modeling achieves similar performance to the variant without any image modeling (Tab.\,\ref{tab:ablation}(a)), with significant decrease compared to Libra with discrete image modeling (\emph{e.g.}, -6.2\% on VizWiz).

\textbf{Impact of Data Scale.}
We reduce the pretraining data size to 3M. In Tab.\,\ref{tab:ablation}(c), we found that larger trainable parameters (4B in Libra) require more training data for convergence.

\textbf{Impact of Routed Visual Expert.}
In Tab.\,\ref{tab:ablation}(d), we remove the routed visual expert design in Libra, where Libra degenerates to LLaVA~\cite{llava}. The results show that the coupled vision and language systems exhibit obvious performance degradation on zero-shot tasks.

We further investigate the impact of the cross-modal bridge in the routed visual expert, as shown in Tab.\ref{tab:ablation}(e). The results show that a simple visual expert alone yields limited performance benefits, \emph{i.e.}, the variant with a simple visual expert (Tab.\,\ref{tab:ablation}(e)) vs. the variant without any visual experts (Tab.\,\ref{tab:ablation}(d)). This indicates the importance of a reasonable cross-modal interaction strategy. Fig.\,\ref{fig:img_modeling}(a) provides further evidence of the effectiveness of the cross-modal bridge.

\textbf{Impact of Hybrid Vision Inputs.}
The contiguous visual signal plays a crucial role in accurate visual perception. To validate this, we remove the contiguous visual signals in Libra's hybrid inputs and only retain  the discrete embeddings. As shown in Tab.\,\ref{tab:ablation}(f), a clear performance degradation rises when solely using discrete inputs (\emph{e.g,}, -20.9\% on VizWiz). 
Tab.\,\ref{tab:ablation}(g) presents the variant without the CLIP encoder.

\section{Conclusion}
Through building up Libra, we found that a decoupled vision system can boost vision-language comprehension in the image-to-text scenario. Libra achieves this through the routed visual expert design, where a simple visual expert ensures separate parameter spaces for vision and language, and a cross-modal  bridge module decouples inner-modal modeling and cross-modal interaction. 
Meanwhile, the hybrid image tokenization enables both contiguous visual comprehension and stable discrete modeling.
We found that the design of Libra yields diverse attention patterns across layers, indicating potentially low learning redundancy.
Vision and language should be integrated in a more reasonable manner beyond simple modality alignment.
We hope our work could provoke more consideration in MLLM designs.

\section*{Acknowledgements}
We'd like to thank Menghao Hu for data management, and Chaoyou Fu  for early discussion.
This work was supported by National Natural Science Foundation of China (No. 62036012, U23A20387, 62322212, 62072455).

\section*{Impact Statement}
Libra presents a range of advantages along with potential risks. Its capability to quickly adapt to diverse tasks has the potential to empower non-expert users to achieve satisfactory performance even in data-scarce scenarios. This characteristic can lower the barriers for beneficial applications, but this flexibility also raises concerns regarding malicious and negative applications, necessitating careful consideration.  Additionally, Libra shares similar risks with LLMs, such as generating offensive language, perpetuating social biases and stereotypes, and potential privacy breaches. Furthermore, Libra's promising capability to process visual inputs introduces specific risks, including gender and racial biases associated with input image content.
To mitigate these risks, we take various measures, such as utilizing debiased pretraining datasets, blurring human faces in training data, and carefully validating instruction tuning data.

\nocite{mqdet,evo-vit,xu2022towards,xu2023exploring,xu2022transformers,yao2024spikedriven,yao2024spike}

\bibliography{reference}
\bibliographystyle{icml2024}

\newpage
\appendix
\onecolumn
\section{Implementation Details}
\subsection{Image Tokenization}
\label{sec:tokenizer_appendix}
\textbf{Tokenization Process.}
Thanks to lookup-free quantization (LFQ)~\cite{lfq}, we can enlarge the vision vocabulary size to $2^{18}$. However, directly implementing such a large vocabulary size brings huge computing costs, \emph{e.g.}, 1B parameters of a simple linear prediction head. Therefore, we predict in two concatenated codebooks, each of size $2^{9}$.

Specifically, the image tokenization process can be illustrated as:
\begin{equation}
\begin{split}
    E_{c}&=\Phi(I), \quad id_{1}, id_{2} = LFQ(E_{c}), \quad E_{d} = \texttt{concat}(E_{1}(id_{1}), E_{2}(id_{2})), \quad X_{I} = \texttt{concat}(E_{c}, E_{d}),\\
\end{split}  
\end{equation}
where $I$ is an input image, $\Phi$ is the image encoder, $E_{c}$ refers to the contiguous visual signals, $E_{1}$ and $E_{2}$ are two separate vision ``word'' embedding banks, $E_{d}$ denotes the discrete vision embeddings, and $X_{I}$ represents for the vision inputs of the Libra model. $\texttt{concat}$ denotes concatenation in the channel dimension.
During LFQ, we utilize two codebooks, each with a size of $2^9$, to predict the tokenized ids $id_{1}$ and $id_{2}$. The sizes of embedding banks $E_{1}$ and $E_{2}$ are also set to $2^9$.
We maintain two prediction heads for vision outputs of Libra to separately predict $id_{1}$ and $id_{2}$. This design largely reduces the computational costs, \emph{i.e.}, the prediction heads comprise: 2 (codebook number) $\times$ 2M (head parameters) = 4M parameters.

\begin{figure}[ht]
    \centering
    \includegraphics[width=0.8\linewidth]{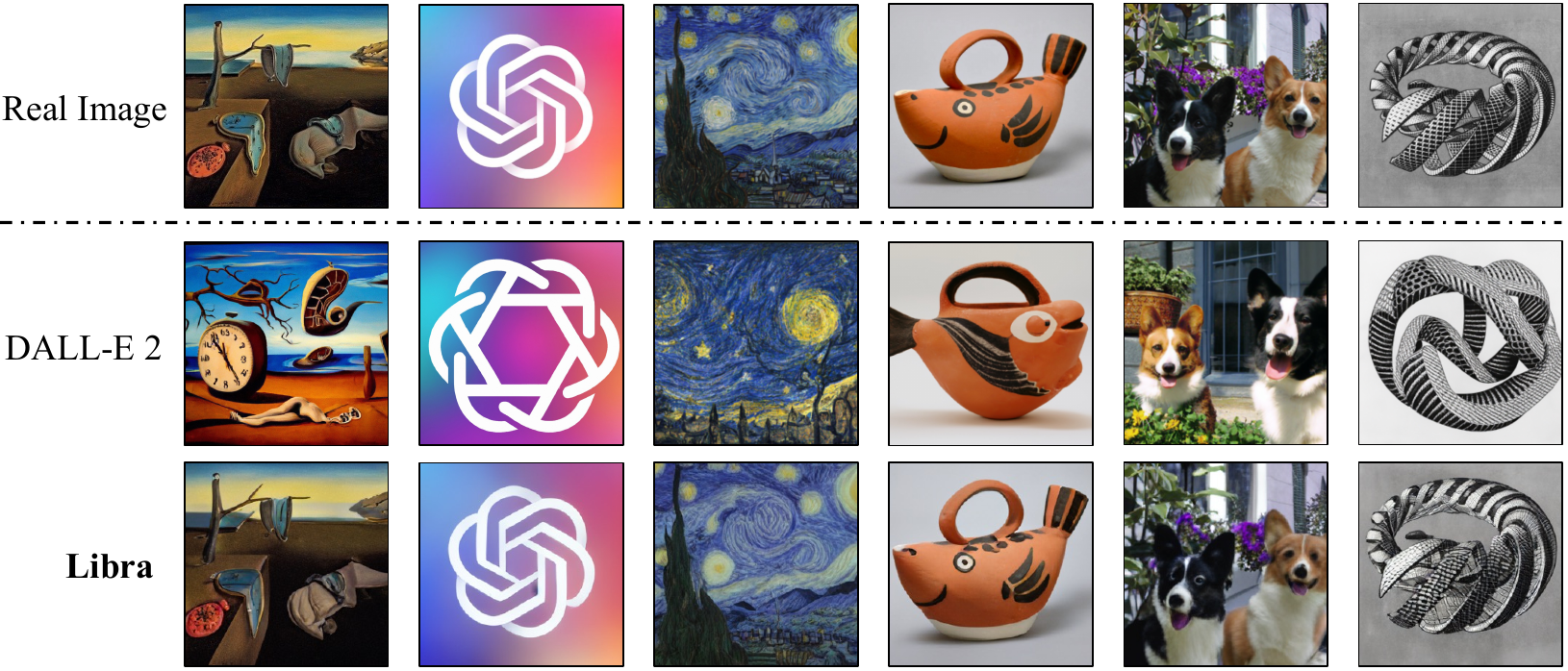}
    \caption{Image reconstruction results of the image tokenizers in Libra and DALL-E 2~\cite{dalle2}.}
    \label{fig:vs_dalle2}
\end{figure}

\textbf{CLIP-based Image Tokenizer.}
Libra's image tokenizer is built on a CLIP-based VQGAN with LFQ. As far as we know, it is the first time that a highly reconstructive image tokenizer can be constructed based on a frozen vision encoder like CLIP~\cite{clip}. The most relevant approach to our image tokenizer is the one in DALL-E 2~\cite{dalle2}, which tokenizes images into discrete embeddings using a frozen CLIP image encoder and decodes to original images with a diffusion~\cite{latent_diffusion} decoder. We compare the reconstruction performance of the tokenizers in Libra and DALL-E 2 in Fig.\,\ref{fig:vs_dalle2}. As the results show, the tokenizer of DALL-E 2 can catch basic visual concepts but largely distort the original visual information. This is helpful for diverse text-to-image generation but detrimental to accurate image-to-text comprehension. In contrast, the tokenizer of Libra effectively captures comprehensive visual information while preserving the pretrained CLIP knowledge.

\textbf{Brief Introduction of LFQ.}
Lookup-free quantization (LFQ)~\cite{lfq} reduces the embedding dimension of the VQ codebook~\cite{vqvae} to zero. Specifically, the codebook $\mathbf{C} \in \mathbb{R}^{K \times d}$, similar to the one in VQGAN~\cite{vqgan}, is replaced with an integer set $\mathbb{C}$ where $|\mathbb{C}|=K$, where $K$ is the vision vocabulary size and $d$ represents the embedding dimension. This approach eliminates the need for embedding lookup entirely. Unlike previous quantization methods~\cite{vqgan,vqvae} that require codebook embeddings to mimic input features for image reconstruction, LFQ does not require such emulation as it has no codebook embeddings. In light of this, we utilize LFQ to successfully quantize highly semantic CLIP features, which has not even been investigated in LFQ.

\subsection{Training Details}
\label{sec:training_appendix}
 \textbf{Training Hyperparameter.}
 The training process of Libra consists of 3 stages: language pretraining (already done in the pretrained LLM), multimodal pretraining, and 
instruction tuning/supervised finetuning (SFT). We present the hyperparameters of Libra during multimodal pretraining and instruction tuning stages in Tab.\,\ref{tab:hyperparam}. The multimodal pretraining stage takes 8400 NVIDIA A100-40G GPU hours and the instruction tuning stage takes 380 NVIDIA A100-40G GPU hours.
\begin{table}[h]
\caption{Training hyperparameters of Libra in different stages.}
\label{tab:hyperparam}
\vskip 0.1in
\begin{center}
\begin{small}
\begin{tabular}{l|ll}
\toprule
Configuration       & Pretraining        & SFT        \\
\midrule
Total steps         & 40000              & 7000       \\
Warmup steps        & 2000               & 300        \\
Batch size          & 1280               & 128        \\
Learning rate       & 1e-4               & 2e-5       \\
Learning rate decay & \multicolumn{2}{c}{cosine decay}      \\
Weight decay        & \multicolumn{2}{c}{0.01}        \\
Dropout ratio       & \multicolumn{2}{c}{0.0}         \\
Optimizer           & \multicolumn{2}{c}{AdamW}       \\
Adam $\epsilon$              & \multicolumn{2}{c}{1e-8}        \\
Adam $\beta$              & \multicolumn{2}{c}{(0.9, 0.99)} \\
Gradient clipping   & \multicolumn{2}{c}{1.0}         \\
Numerical precision & \multicolumn{2}{c}{bfloat16}    \\ 
\midrule
LLM                 & \multicolumn{2}{c}{LLaMA2-7B-Chat}        \\
Vision encoder      & \multicolumn{2}{c}{CLIP-ViT-L-336px}        \\
Image resolution    & \multicolumn{2}{c}{336$^{2}$}         \\
Patch size          & \multicolumn{2}{c}{14 $\times$ 14}         \\
Image token number    &  \multicolumn{2}{c}{578}              \\
Vision vocab size   & \multicolumn{2}{c}{$(2^{9})^{2} = 2^{18}$} \\
Language vocab size & \multicolumn{2}{c}{32000} \\
\bottomrule
\end{tabular}
\end{small}
\end{center}
\vskip -0.1in
\end{table}

\textbf{Instruction Template.}
We arrange the instruction tuning data based on the template described in Sec\,\ref{sec:training}. The \texttt{<System Message>} in Eqn.\,(\ref{eqn:templete}) is:

\begin{equation}
    \begin{split}
        &\prompt{\text{A chat between a curious user and an artificial intelligence assistant.}}\\
        &\prompt{\text{The assistant gives helpful, detailed, and polite answers to the user's questions.}}
    \end{split}
\end{equation}

\section{Evaluation Details}
\label{sec:eval_details_appendix}
\subsection{Benchmarks and Metrics}
\label{sec:benchmark_appendix}
\begin{table*}[t]
\caption{Detailed information of the evaluation benchmarks.}
\label{tab:metric}
\vskip 0.1in
\begin{center}
\begin{small}
\resizebox{\linewidth}{!}{
\begin{tabular}{c|l|lll}
\toprule
Task & Dataset & Description & Split & Metric \\
\midrule
\multirow{5}{*}{General VQA} 
& VQAv2 & VQA on natural images. 
& test-dev  & VQA Score ($\uparrow$) \\

& OKVQA & VQA on natural images requiring outside knowledge.  
& val       & VQA Score ($\uparrow$)  \\

& GQA   & VQA on scene understanding and reasoning.           
& test-balanced  & EM ($\uparrow$)  \\

& VizWiz  & VQA on photos taken by people who are blind.      
& test-dev   &  VQA Score ($\uparrow$) \\

& SQA     & Multi-choice VQA on a diverse set of science topics. 
& Img-test  & Accuracy ($\uparrow$)  \\

\midrule
\multirow{3}{*}{Image Caption} 
& NoCaps & Captioning of natural images.   
&  val    & CIDEr ($\uparrow$) \\

& Flickr  & Captioning of natural images.  
& karpathy-test    & CIDEr ($\uparrow$) \\

& COCO    & Captioning of natural images.  
& karpathy-test   & CIDEr ($\uparrow$) \\

\midrule
\multirow{8}{*}{MLLM Benchmark}  

& POPE  & Object existence by yes/no questions.  
& random/popular/adversarial  &  F1 Score($\uparrow$) \\

& MME   & Visual perception by yes/no questions.   
& Perception   & MME Score ($\uparrow$)  \\

& MME$^{C}$  & Visual cognition by yes/no questions.  
& Cognition   & MME Score ($\uparrow$)  \\

& MMB  & Multi-choice VQA with circular evaluation. 
& test   &  Accuracy ($\uparrow$) \\

& MMB$^{CN}$  & Multi-choice VQA in Chinese with circular evaluation.  
&  test  &  Accuracy ($\uparrow$) \\

& SEED  &  Open-ended multi-choice VQA.              
& Image \& Video   &  Accuracy ($\uparrow$) \\

& MM-Vet & Open-ended VL benchmark with various abilities                          &  test  &  GPT-4 Score ($\uparrow$) \\

& MMVP & Detecting CLIP bias by multi-choice VQA.          &  test  &  GPT-4 Score ($\uparrow$) \\

\bottomrule
\end{tabular}
}
\end{small}
\end{center}
\vskip -0.1in
\end{table*}
We provide detailed information of the evaluation benchmarks used in this work in Tab.\,\ref{tab:metric}.
We use different language prompts for each dataset according to corresponding data forms, as shown in Tab.\,\ref{tab:prompt}.

\subsection{Details on Attention Difference}
\label{sec:attn_diff}
\begin{figure}[t]
    \vskip 0.2in
    \centering
    \includegraphics[width=0.9\linewidth]{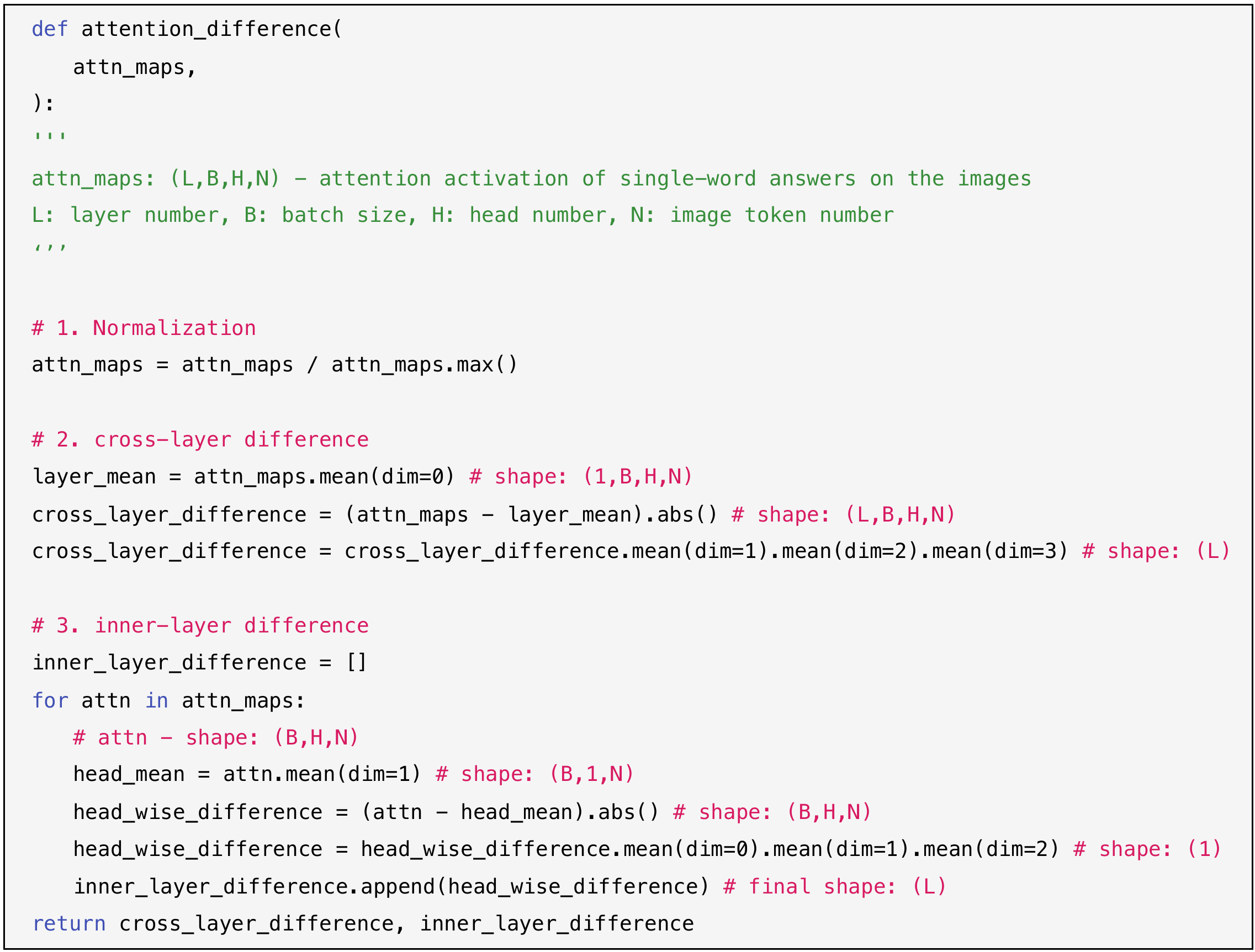}
    \caption{Pseudo code for the computing process of attention differences in Fig.\,\ref{fig:vis_attn}.}
    \label{fig:pseudo}
\end{figure}

We provide more details on the computing process of attention differences in Fig.\,\ref{fig:vis_attn}. We present the pseudo code in Fig.\,\ref{fig:pseudo}.

\subsection{Comparison in the Era of Foundation Models}
In the era of foundation models, it is hard to achieve a completely fair performance comparison due to numerous variables such as model parameter size, model architecture, and training data. It is only possible to conduct relatively fair comparisons in scenarios where these variables are approximately equal. Meanwhile, smaller model parameters often imply easier training, which can lead to better performance particularly when dealing with limited data. For example, LLaVA1.5~\cite{llava} achieves remarkable performance with merely 1.2M total training data, thanks to its small trainable parameter size (30M during pretraining).
We perform the comparison between Libra and other MLLMs under similar model parameter sizes.
Last but not least, \textit{Libra is not intended to achieve state-of-the-art performance; rather, it serves as a prototype model.} Our aim in developing Libra is to offer a promising perspective beyond simple modality alignment for the design of future MLLMs. The evaluations conducted on the Libra prototype have effectively showcased the potential of the decoupled vision systems within MLLMs.

\section{Qualitative Evaluation}
\begin{figure}[t]
    \vskip 0.2in
    \centering
    \includegraphics[width=\linewidth]{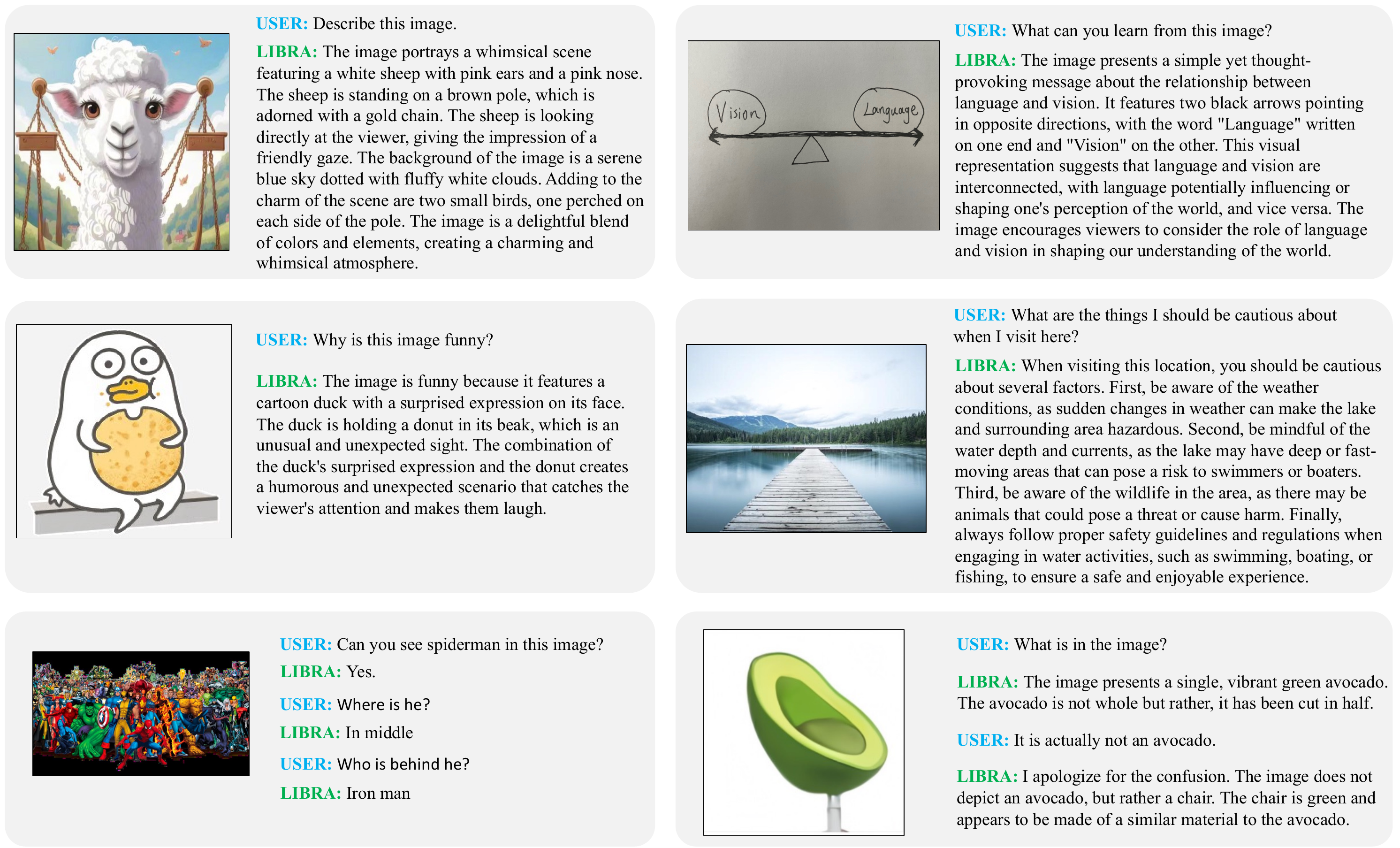}
    \caption{Conversations between users and Libra.}
    \label{fig:chat}
\end{figure}

Fig.\,\ref{fig:chat} shows several conversations between users and Libra. We discovered that Libra demonstrates robust visual perception capabilities and inherits the cognitive abilities of LLMs. For example, it is capable of identifying objects within an image and performing further deducing (\emph{e.g.}, the funny point of an image). Additionally, Libra can catch the relationship between objects, \emph{e.g.}, locations. We also found that Libra, like many commercial chatbots, is capable of error correction based on the user feedback, as shown in the last case.

\section{Further Discussion}
\subsection{Societal Impact}
In terms of societal impact, Libra presents a range of advantages along with potential risks. Its capability to quickly adapt to diverse tasks has the potential to empower non-expert users to achieve satisfactory performance even in data-scarce scenarios. This characteristic can lower the barriers for beneficial applications, but this flexibility also raises concerns regarding malicious and negative applications, necessitating careful consideration.  Additionally, Libra shares similar risks with LLMs, such as generating offensive language, perpetuating social biases and stereotypes, and potential privacy breaches. Furthermore, Libra's promising capability to process visual inputs introduces specific risks, including gender and racial biases associated with input image content.
To mitigate these risks, we take various measures, such as utilizing debiased pretraining datasets, blurring human faces in training data, and carefully validating instruction tuning data.

\subsection{Limitations}
First, our model is built on pretrained LLMs, and as a side effect, directly inherit their weakness. For example, LLM priors generally provide helpful contextual information, but occasionally demonstrate hallucinations and ungrounded guesses.
In addition, it is observed that LLMs exhibit poor generalization when faced with sequences longer than the ones they were trained on.

Second, the routed visual expert design introduces a novel attention computing mechanism, which has not been officially supported by existing acceleration frameworks (\emph{e.g.}, FlashAttention~\cite{flashattention}) yet. Addressing this issue can make Libra more efficient and more friendly to the downstream implementation.

\begin{table}[t]
\caption{Language prompts for different datasets.}
\label{tab:prompt}
\vskip 0.1in
\begin{center}
\begin{small}
\resizebox{\linewidth}{!}{
\begin{tabular}{l|l|p{10cm}}
\toprule
Task           & Dataset     & Language Prompt \\
\midrule
\multirow{5}{*}{General VQA}     
& VQAv2~\cite{vqa}       
& Answer the question using a single word or phrase.               \\
& OKVQA~\cite{okvqa}       
& Answer the question using a single word or phrase.               \\
& GQA~\cite{gqa}         
& Answer the question using a single word or phrase.               \\
& VizWiz~\cite{vizwiz}      
& When the provided information is insufficient, respond with `Unanswerable'. Answer the question using a single word or phrase.  \\
& SQA~\cite{sqa}         
& Answer with the option's letter from the given choices directly. \\
\midrule
\multirow{3}{*}{Image Caption} 
& NoCaps~\cite{nocaps}      
& Provide a one-sentence caption for the provided image.           \\
& Flickr~\cite{flickr30k}      
& Provide a one-sentence caption for the provided image.           \\
& COCO~\cite{cococap}        
& Provide a one-sentence caption for the provided image.           \\
\midrule
\multirow{8}{*}{MLLM Benchmark} 
& POPE~\cite{pope}        
& Answer the question using a single word or phrase.   \\
& MME~\cite{mme}              
& Answer the question using a single word or phrase.   \\
& MME$^{C}$~\cite{mme} 
& Answer the question using a single word or phrase.   \\
& MMB~\cite{mmb}         
& Answer with the option's letter from the given choices directly.  \\
& MMB$^{CN}$~\cite{mmb}       
& \begin{CJK}{UTF8}{gbsn} 请直接回答选项字母。\end{CJK} \\
& SEED~\cite{seed}        
& Answer with the option's letter from the given choices directly.  \\
& MM-Vet~\cite{mm-vet}      
& -                                                                 \\
& MMVP~\cite{mmvp}        
& Answer this question concisely.          \\
\bottomrule
\end{tabular}
}
\end{small}
\end{center}
\vskip -0.1in
\end{table}

\end{document}